\documentclass{article}

\usepackage{microtype}
\usepackage{graphicx}
\usepackage{subfigure}
\usepackage{booktabs}

\usepackage{arxiv}

\usepackage[utf8]{inputenc} 
\usepackage[T1]{fontenc}  
\usepackage{url}         
\usepackage{amsfonts}    
\usepackage{nicefrac}     
\usepackage{lipsum}
\usepackage{graphicx}
\graphicspath{ {./images/} }

\usepackage{times}
\usepackage{epsfig}
\usepackage{graphicx}
\usepackage{float}
\usepackage{wrapfig}
\usepackage{amsmath,amssymb,amsthm}
\usepackage{bm,xspace}
\usepackage{comment}
\usepackage{verbatim}
\usepackage{multirow}
\usepackage{balance}
\usepackage{url}
\usepackage{booktabs}
\usepackage{etoolbox}
\usepackage{calc}
\usepackage{pifont,hologo}

\newcommand{\ra}[1]{\renewcommand{\arraystretch}{#1}}

\usepackage[super]{nth}
\usepackage{nicefrac}
\robustify\bfseries

\def\aa{\mathbf{a}}

\DeclareMathSymbol{@}{\mathord}{letters}{"3B}

\def\latex/{\LaTeX}
\def\bibtex/{\hologo{BibTeX}}

\usepackage{xcolor}

\usepackage{mathtools}
\usepackage{bbm}

\usepackage{natbib}
\setcitestyle{authoryear,round,citesep={;},aysep={,},yysep={;}}

\newcommand{\beginsupplement}{%
  \setcounter{table}{0}
  \renewcommand{\thetable}{S\arabic{table}}%
  \setcounter{figure}{0}
  \renewcommand{\thefigure}{S\arabic{figure}}%
}

\renewcommand{\cite}[1]{\citep{#1}}

\makeatletter
\newcommand{\printfnsymbol}[1]{%
  \textsuperscript{\@fnsymbol{#1}}%
}
\makeatother

\title{Scaling Imitation Learning in Minecraft}

\usepackage{authblk}

\author[1]{\textbf{Artemij Amiranashvili}\thanks{equal contributors}\hspace{1.6mm}}
\author[1]{\textbf{Nicolai Dorka}\printfnsymbol{1}}
\author[1]{\textbf{Wolfram Burgard}}
\author[2]{\textbf{Vladlen Koltun}}
\author[1]{\textbf{Thomas Brox}}

\affil[1]{University of Freiburg}
\affil[2]{Intel Labs}

\usepackage{hyperref}

\begin{document}

\maketitle

\begin{abstract}
Imitation learning is a powerful family of techniques for learning sensorimotor coordination in immersive environments. We apply imitation learning to attain state-of-the-art performance on hard exploration problems in the Minecraft environment. We report experiments that highlight the influence of network architecture, loss function, and data augmentation. An early version of our approach reached second place in the MineRL competition at NeurIPS 2019. Here we report stronger results that can be used as a starting point for future competition entries and related research. Our code is available at \url{https://github.com/amiranas/minerl_imitation_learning}.
\end{abstract}


\section{Introduction}
\label{intro}

Reinforcement learning (RL) was used to reach outstanding performance in many challenging domains, such as Atari~\citep{mnih2015human}, Go~\citep{silver2017mastering}, Starcraft II~\citep{vinyals2019grandmaster}, and immersive 3D environments~\citep{jaderberg2019human,ddppo,petrenko2020sf}. However, RL algorithms require billions of interaction steps with the environment in order to achieve these results. This makes application of RL challenging for slow environments that cannot be run for billions of time steps with available computational resources.

In RL training, the agent needs to encounter rewards in the environment before policy optimization can begin. Therefore, environments with very sparse rewards and large action spaces, such as Minecraft, require a huge amount of exploration, exacerbating the sample inefficiency of RL~\cite{guss2019minerl}. 

Another approach to obtain a policy for challenging domains is to use imitation learning, which trains directly from expert demonstrations. It does not require any interaction with the environment during training and is unhindered by the sparsity of rewards, making it a desirable alternative to RL. We evaluate the performance of imitation learning on complex Minecraft tasks, for which a large amount of demonstration data is available.

Minecraft is a first-person open world game where the agent interacts with a procedurally-generated 3D environment. We focus on the \textit{ObtainIronPickaxe} task, which consists of 11 distinct subtasks and features very sparse rewards. A large-scale expert demonstration dataset has been made available for this task by~\citet{guss2019minerl}, making it a suitable and challenging testbed for imitation learning.

We first introduce state- and action-space representations together with demonstration data processing that enables successful imitation learning in Minecraft. We use this setup to investigate how factors such as network architecture, data augmentation, and loss function affect imitation learning performance.

We applied an early form of the presented approach in the MineRL competition at NeurIPS 2019~\citep{guss10minerlcompetition}, which deliberately constrained computational resources and environment interactions. Our entry reached the second place in the competition without using the environment during training~\citep{milani2020retrospective}. Here we present a stronger form of our approach that attains higher performance and can be used as a starting point for future competitions and related research.

\section{Minecraft Environment}
\label{minecraft}

In Minecraft the player interacts with a procedurally generated, 3D, open-world environment. 
According to the specified task, different goals must be achieved, such as finding and gathering important resources or using the obtained resources to craft better tools required for getting access to more resources. 
The agent observes the world from a first-person perspective and also has information about the obtained resources, making it a partially observable environment. A simulator Malmo has been created by \citet{johnson2016malmo} to support research on the Minecraft domain. Previous works have used the environment to tackle navigation problems~\cite{matiisen2019teacher} or block stacking tasks~\citep{shu2017hierarchical}. Malmo has also been used by \citet{guss2019minerl} to create an array of defined scenarios within the Minecraft environment, such as the \textit{ObtainIronPickaxe} task. 
Additionally, \citet{guss2019minerl} released the \textbf{\emph{MineRL-v0}} demonstration dataset, with over 500 hours of human trajectories of the introduced tasks. 
This is the first time a large-scale demonstration dataset has been provided for an image-based environment, which makes it a suitable testbed for imitation learning approaches.

\subsection{Competition}
The MineRL  competition at NeurIPS 2019 used the Minecraft environment as a testbed~\citep{guss10minerlcompetition}.
The goal was to solve the \textit{ObtainDiamond} task, where an agent has to fulfill many subtasks in order to reach a diamond. 
To do so the agent was allowed to learn from the \textbf{\emph{MineRL-v0}} dataset and was further allowed 8 million interaction steps with the Minecraft environment. In the final round the agent had to be trained remotely on a single machine with a single GPU in 4 days. Custom environment textures that are not public were used during the training and evaluation. The restricted compute power, training time, and interactions with the environment made it an interesting and challenging setting for imitation learning.

\begin{figure*}[t!]
\begin{minipage}{\linewidth}
\centering
\includegraphics[scale=0.6]{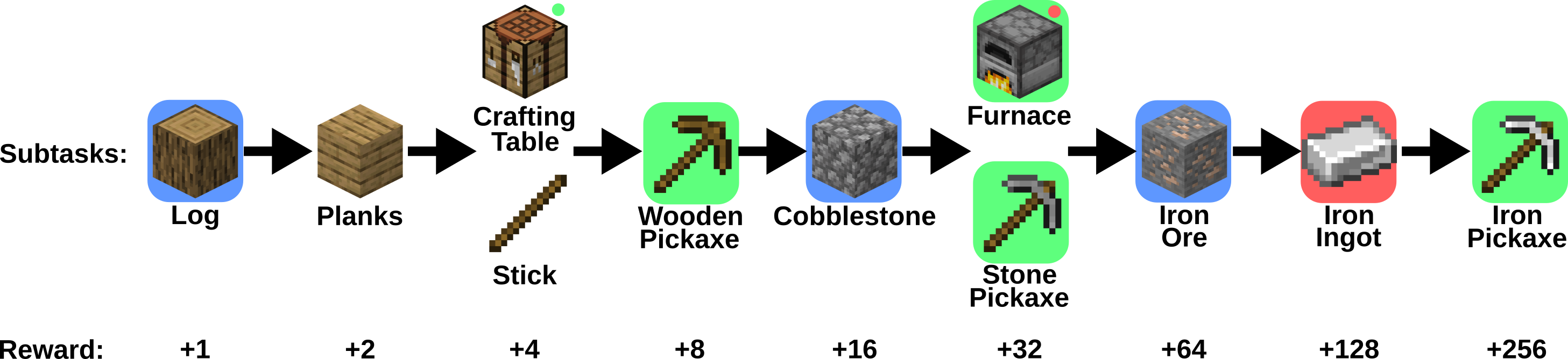}
\end{minipage}
\caption{Subtasks of the \textit{ObtainIronPickaxe} scenario with the respective rewards. Items with blue background are resources that have to be found and gathered in the Minecraft world. To build the items with a green background a crafting table is required and a furnace is required to smelt iron ores into ingots. Cobblestone can only be obtained with a pickaxe and iron ore only with a stone pickaxe. Each reward is only given once per item type. The sequential dependency of the subtasks makes learning a robust policy difficult, since failure in a single step prevents the policy from progressing further.}
\label{fig:irontask}
\end{figure*}

\subsection{Minecraft Tasks}

We focus on two Minecraft tasks: \textit{ObtainIronPickaxe} and \textit{Treechop}.

\paragraph{ObtainIronPickaxe.} In this task the agent has to fulfill a sequence of 11 subtasks in order to craft an iron pickaxe. The subtasks are divided into two categories. One is gathering different resources like wood, stone, and iron. The other subtasks are using the collected resources to create tools like pickaxes. With better tools the agent is able to gather better resources and build better pickaxes. Some of the crafting subtasks require a special tool, like the crafting table, which must be placed in front of the agent. The full sequence of subtasks and the associated rewards are shown in Figure \ref{fig:irontask}. The task is identical to the \textit{ObtainDiamond} task from the MineRL competition at NeurIPS 2019, except that the last subtask of obtaining a diamond with the iron pickaxe has been removed. We removed the last step because so far none of the tested methods or competition entries were able to obtain a diamond. We use the same maximum episode length as in the \textit{ObtainDiamond} task to have the same episode length constraint as in the competition.

\paragraph{Treechop.} For this task, the agent always starts in the forest biome where enough trees are present and has to navigate through the world, find trees and ``attack'' them to obtain logs of wood. For each log obtained in this manner the agent receives a reward of 1. The task is considered solved once 64 logs are collected. This task is only used for a comparison with reinforcement learning in Section~\ref{crl}.

\section{Methods}

\subsection{Imitation Learning}

\label{imitation}

We work with environments that are discretized in timesteps $t$. At every step the agent receives an observation of the environment $s_t$ and has to choose an action $\aa_t$. 
Thereafter the environment is progressed by one timestep and the agent receives feedback in the form of a reward $r_t$ and the next observation $s_{t+1}$, until it reaches a terminal state. The objective is to find a policy $\pi(s)$ that collects the highest return over an episode: $R = \sum_{t=0}^{T} r_{t}$. 

Imitation learning aims to maximize the return by imitating the behavior of human trajectories. Usually a network is trained to predict an action given an observation on the demonstration data. This makes imitation learning a classification problem, with the different actions as classes. 

Imitation learning has been successfully applied to other domains, such as Atari games \citep{bogdanovic2015deep}, Mario \citep{chen2017game}, or Starcraft 2 \citep{vinyals2019grandmaster, justesen2017learning}.
We evaluate different approaches to train an imitation learning policy. First we consider a classification-based approach with a policy defined through a neural network with a softmax activation function after the last layer:
\begin{equation}
\begin{aligned}
\pi(s, a) = p_\pi(a | s) & = \text{Softmax}_a(f(s, a)),
\end{aligned}
\end{equation}
where $f(s,\cdot)$ denotes the features of the last layer and has a length equal to the amount of possible actions. We train the policy $\pi(s, a)$ to predict the expert action through a cross-entropy loss. Thereafter the action is either sampled from the distribution ($a \sim p_\pi(a|s) $) or the action with the highest probability is selected ($a = \text{argmax}_a p_\pi(a|s) $). This supervised learning based policy training is often referred to as Behavior Cloning.

We also evaluate the performance of the pre-training process of Deep Q-learning from Demonstrations (DQfD) \citep{hester2018deep}. There the reward signal is incorporated into the training process together with imitation learning. In their case a greedy policy is used, which selects the action with the highest action-value: $\pi(s) = \text{argmax}_a Q(s,a)$. The action-value function is trained with the Q-learning loss~\citep{mnih2015human}:
\begin{equation}
Q^\text{Target}(s, a) = R_n + \gamma^n \max_a Q(s', a).
\end{equation}
The expert action information is incorporated through an additional margin loss. 
In a state $s$, let $a_E$ be the action of the expert and $m(a_E, a) \coloneqq b \cdot \mathbbm{1}\{a\ne a_E\}$ a margin function that is $0$ when $a=a_E$ and takes a value $b>0$ otherwise. The large-margin classification loss \citep{piot2014marginloss} is defined as
\begin{equation}
L_S = \max_{a \in A}\big[Q(s,a) + m(a_E, a) \big] - Q(s, a_E).
\end{equation}
The margin loss is minimized if $a_E$ is the maximizer of $Q(s, \cdot)$ and if its value at $a_E$ exceeds that of any other action $a$ by at least the margin $b$. As a result, minimizing this loss pushes the Q-function to assign higher values to the expert actions. In the experiments we compare the empirical performance of the margin loss, with and without the TD loss, to that of the cross-entropy loss. Without the reward incorporation the margin classification loss becomes
\begin{equation}
L_S = \max_{a \in A}\big[f(s,a) + m(a_E, a) \big] - f(s, a_E).
\end{equation}

\subsection{Training Setup}
\label{sec:setup}

In this section we describe the training details of applying imitation learning to the Minecraft domain and the considered alterations.
We want to investigate how much the training setup, such as the choice of the network architecture or the use of data augmentations, influences the performance of imitation learning. 

\paragraph{State and Action Space.}
The main part of the state is the observation of the environment that consists of a $64 \times 64$ RGB image. In the \textit{ObtainIronPickaxe} task an additional vectorial state is provided that consists of information about the collected resources and crafted tools, and the currently held item. We encode the held item as a one-hot vector and additionally encode the items in the inventory as multi-hot vectors (amount of ones in a sub-vector equals the amount of the according item in the inventory).

The action space consists of three parts. First there are 8 binary actions related to the movement in the environment: \textit{forward}, \textit{backward}, \textit{left}, \textit{right}, \textit{jump}, \textit{sprint}, \textit{attack}, \textit{sneak}. Multiple movement actions can be used in the same timestep, resulting in 256 combinations. The second part is the continuous yaw and pitch control of the agent's camera orientation. The last part are the actions related to crafting, equipping, and placing of items. Some items, like the crafting table, require being placed on the ground before they can be used.

The combination of these basic actions results in a massive action space. We implement continuous control by quantization, which is a common practice for first-person environments~\citep{jaderberg2019human, kempka2016vizdoom}. We choose a single rotation value of 22.5 degrees for each direction. 

After quantization of the camera movement, there are 1280 possible movement action combinations. We allow only up to 3 simultaneous movement actions and remove redundant actions like turning left and right at the same time. In the end 112 different movement actions remain. A full description of the movement actions and the state encoding is available in the supplementary material.

\begin{figure}[t!] 
\centering 
\includegraphics[scale=0.43]{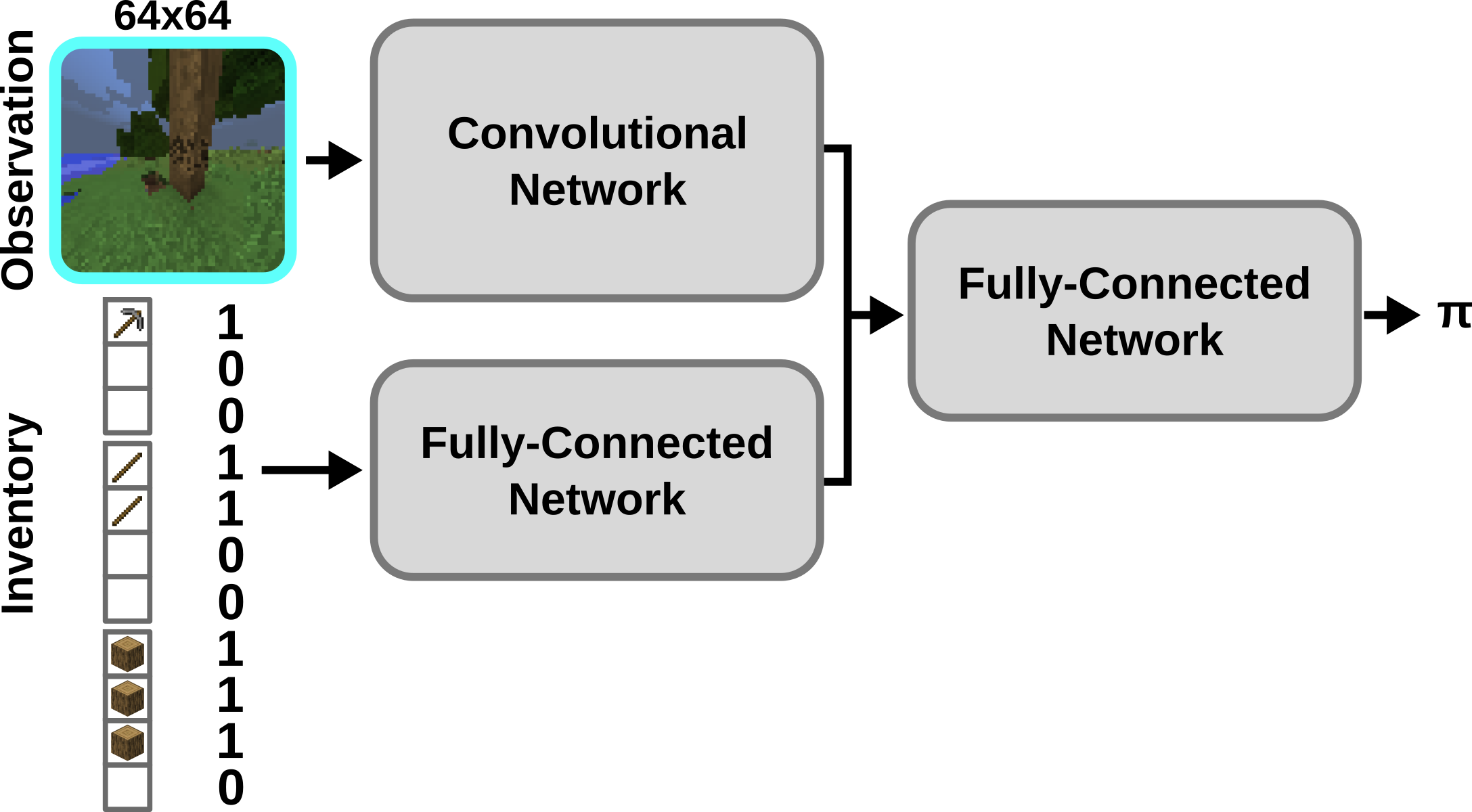} 
\caption{Layout of network architecture and a simplified illustration of the inventory information encoding.}
\label{fig:net}
\end{figure}

\paragraph{Training Setup.}
The policy neural network consists of three parts. 
A convolutional perceptual part for the image input and a fully-connected part for the vectorial part of the state are concatenated after the last layer and followed by a subseqent fully-connected part (Figure \ref{fig:net}). 
The last layer has either a softmax or a linear output for the cross-entropy or the margin loss, respectively. 
We investigate how the network size correlates with performance by testing three different architectures for the perceptual part of the network: the DQN architecture with 3 convolutional layers \citep{mnih2015human}, the Impala architecture with 6 residual blocks~\citep{espeholt2018impala}, and the Deep Impala architecture with 8 residual blocks and Fixup Initialization~\citep{Obstacle,zhang2019fixup}. 
In the last tested architecture we double the channel size of all fully-connected and convolutional layers (Double Deep Impala).

We train the networks with the Adam optimizer, a learning rate of $6.25 \times 10^{-5}$, and weight decay of $10^{-5}$ for up to $3 \times 10^6$ steps. 

From the demonstration dataset we use the human trajectories that successfully reach the target of the environment within the timestep limit of the respective task (\textit{ObtainIronPickaxe} and \textit{ObtainDiamond}). We also remove all states where the human did not perform any action. For full network architectures see the supplementary material. 

We also test multiple augmentations such as horizontal flipping where also left and right actions are flipped, rectangle-removal, brightness, sharpness, contrast, and posterization adjustments.

Beside evaluating the policy performance, we tested two additional measures of performance, the training loss and the test loss on unseen human trajectories, in order to evaluate the correlation of those losses with the actual performance of the policy.

\paragraph{Additional Data Incorporation.}
In the default training setup we used the human trajectories from the \textit{ObtainIronPickaxe} and \textit{ObtainDiamond} tasks. The amount of available training data could be increased by also incorporating the trajectories from the \textit{Treechop} task (where the agent has to collect logs, which is also the first step of the \textit{ObtainIronPickaxe} and \textit{ObtainDiamond} tasks). However, the observation space of the \textit{Treechop} trajectories consists only of the RGB images and no inventory information is available. This makes it incompatible with the trajectories from the other tasks. To create realistic observations for the additional data we first sample a random state from the \textit{ObtainIronPickaxe} and \textit{ObtainDiamond} trajectories, where the reward of $2$ has not yet been reached. Then we use the vectorial observation part of the sampled state to complete a \textit{Treechop} observation. This process is repeated until all \textit{Treechop} states have a complete observation.
\section{Results}

\begin{figure*}[t!]
\begin{minipage}{\linewidth}
\centering
\begin{tabular}{@{}c@{ }c@{ }c@{ }}
 \includegraphics[scale=0.45]{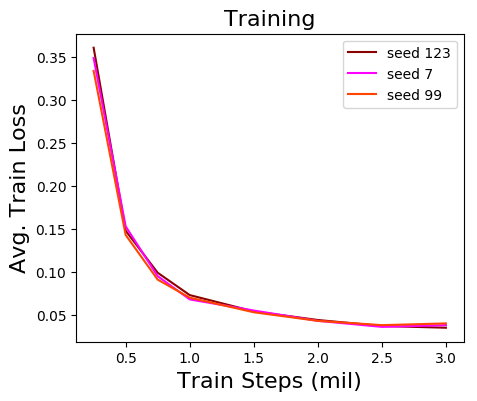}& 
 \includegraphics[scale=0.45]{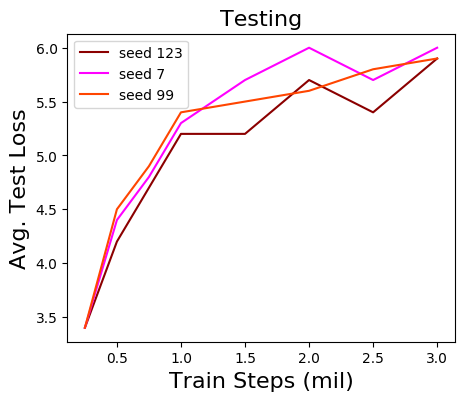}& 
 \includegraphics[scale=0.45]{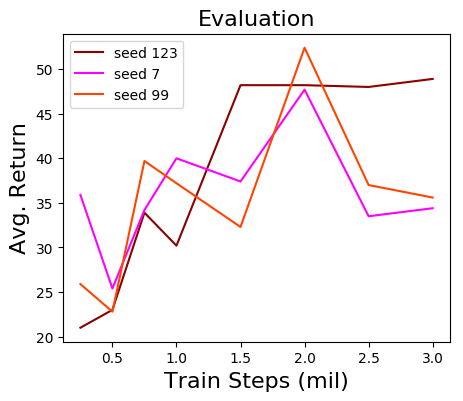}
\end{tabular}
\end{minipage}
\caption{Comparison between the training loss, the test loss, and the obtained average return. The test loss plot shows the cross-entropy loss on human trajectories that were not used during training. The increasing test loss indicates that the human policies are too diverse to be used for reliable predictions across trajectories. Neither of the losses is highly correlated with the actual performance change during training (right figure).}
\label{fig:eval}
\end{figure*}

\subsection{Evaluation} Figure~\ref{fig:eval} shows the training and the test losses across three training runs on the \textit{ObtainIronPickaxe} task. The test loss is the cross-entropy loss of the policy network evaluated on unseen human trajectories. The figure reveals a clear difference between imitation learning and normal supervised training classification: the test loss increases during training, usually a clear indication of heavy overfitting, yet the policy performance in terms of reward keeps improving. Also, even though the training and the test loss were nearly identical between different random seeds, the performance of the snapshots varies over time and is not correlated to either of the losses. Therefore, even when interaction with the environment during training is not required, this interaction cannot be avoided when it comes to evaluation.

We evaluated imitation learning performance as follows. During training 8 snapshots of the network were saved. For each snapshot the average reward was computed from 100 episodes. The performance of the best snapshot was used as the score of the training run. Each training is repeated three times and the average return across the three training runs is considered the overall result of that configuration. The variance was calculated over the three scores of the three training runs. 

In the following plots of this section each point represents the score of the best performing snapshot until that time point, always averaged across three training runs. The shaded area always shows the standard deviation across the three training runs.

\subsection{Architecture and Augmentations}

The performance in terms of architectures and image augmentation options is shown in Table \ref{tab:augment} (all trained with horizontal image flipping augmentations). 
Larger networks yielded better performance. The Deep Impala architecture improved the performance by 78\% and the Double Deep Impala architecture yielded another 7\% improvement.

Image flipping was the only effective augmentation and it improved the performance by $73\%$. Applying additional augmentations had no significant impact and sometimes reduced performance.

\subsection{Margin Loss and Treechop Data Incorporation}

\begin{figure*}[b!]
\begin{minipage}{\linewidth}
\centering
\begin{tabular}{@{}c@{ }c@{ }}
\includegraphics[scale=0.45]{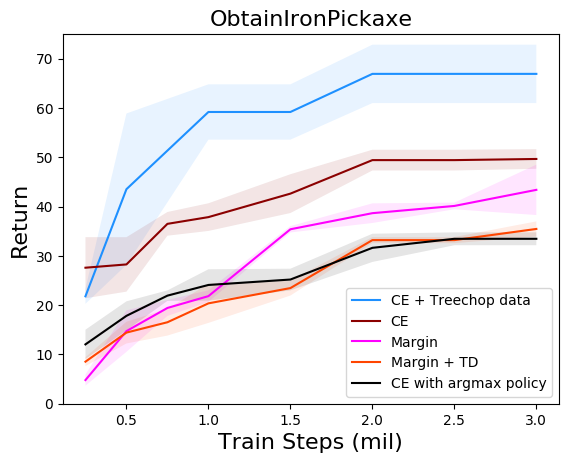}&
\includegraphics[scale=0.45]{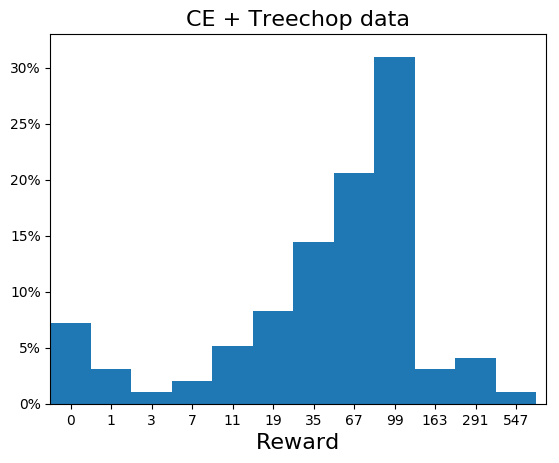}
\end{tabular}
\end{minipage}
\caption{The left figure shows the comparison between different loss functions. The softmax policy trained with the cross-entropy loss (CE) outperformed the deterministic policy based on the margin loss (as used in the DQfD algorithm). However, when an argmax policy was used, instead of sampling, the performance dropped below the deterministic margin policy. The combination of the margin loss with the TD loss only decreased the performance. The plots show the average returns of the best snapshots, averaged across three training runs. The shaded areas show the standard deviation between the three training runs. The right figure shows the reward distributions of the best CE-based policy with additional Treechop data across 100 random seeds.}
\label{fig:td}
\end{figure*}

We compare the cross-entropy loss against the margin loss in Figure~\ref{fig:td} and investigate whether the reward incorporation by the additional TD loss improves the performance. The cross-entropy loss outperformed the margin loss on both tasks. However, it was very important to sample the actions from the softmax distribution of the cross-entropy based policy. When we applied a deterministic argmax policy, the performance of the cross-entropy based policy became worse than the margin loss based policy. This is relevant in cases where a deterministic policy is required. The combination of the margin and TD loss, as used in the DQfD algorithm pretraining phase, diminished performance.

The additional Treechop data improved the performance of the agent by a large margin. This shows that even with the massive human dataset the amount of available trajectories is still a potential bottleneck for the imitation learning approach.

The best policy (Deep Impala agent trained with the cross-entropy loss and additional Treechop data) was able to reach the stone pickaxe in $82\%$ of the episodes. Sometimes it could obtain an iron ingot. An iron pickaxe was only built in rare cases (ca. $1\%$). Typical failure cases included getting stuck in biomes without trees or being buried underground without sufficient resources to finish the task. A histogram of the attained returns is shown in Figure~\ref{fig:td}.

\begin{table}
\centering
\caption{Comparison between architectures and augmentations.}
\ra{1.1}

	\begin{tabular}{@{}l@{\hspace{6mm}}r@{}}
		\toprule
		Network \& Augment. & ObtainIronPickaxe\\
		\midrule
		DQN & $27.9 \pm 2.3$\\
		Impala & $39.2 \pm 5.2$ \\
		Deep Impala & $49.7 \pm 2.0$ \\
		Double Deep Impala & $\mathbf{53.4} \pm 7.5$\\
		\midrule
		Deep Impala no aug. & $28.8 \pm 8.4$ \\
		Deep Impala flipping & $49.7 \pm 2.0$ \\
		Deep Impala flipping + Brightness + rectangle removal & $39.9 \pm 3.0$ \\
		Deep Impala flipping + Sharpness & $50.8 \pm 6.5$ \\
		Deep Impala flipping + Contrast & $47.8 \pm 1.1$ \\
		Deep Impala flipping + Posterization & $42.0 \pm 6.7$ \\
		\bottomrule
	\end{tabular}
\label{tab:augment}
\end{table}

\subsection{Comparison to Reinforcement Learning}
\label{crl}

\begin{figure}[h!]
\centering
\includegraphics[scale=0.54]{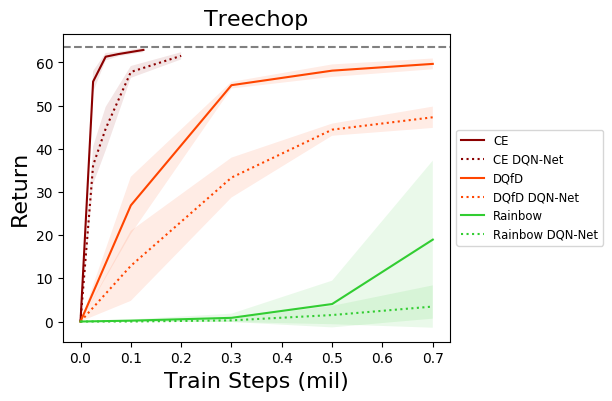}
\caption{Comparison between imitation learning (CE), reinforcement learning (Rainbow) and a combination of the two (DQfD) on the \textit{Treechop} task. The dashed grey line indicates the maximal possible return of the task.}
\label{fig:RL}
\end{figure}

For the action space used in this work, the only task (out of the \textit{Treechop}, \textit{ObtainIronPickaxe}, and  \textit{ObtainDiamond} tasks) where Rainbow \citep{hessel2018rainbow}, an RL approach without use of demonstrations, was able to outperform a random policy was the \textit{Treechop} task. We compare the results of Rainbow to imitation learning with the cross-entropy loss and the DQfD algorithm in Figure~\ref{fig:RL}. For all methods we show a variant with the DQN and the Deep Impala network architecture. A larger network always improved results. Imitation learning was able to reach near-optimal performance after just $50000$ train steps. 
The RL approach was able to obtain non-zero rewards only on two of the three training runs.

\section{MineRL Competition Results}

A early form of the presented imitation learning approach placed 2nd in the MineRL competition at NeurIPS 2019, out of 41 participating teams, and received the ``Surprising Research Result'' award (for attaining this level of performance without using the Minecraft environment at all during training)~\citep{milani2020retrospective}. The final round was rated by the best-performing policy out of four separate training runs. Our best policy at the time yielded an average return of $42.4$, while the top entry in the competition achieved an average return of $61.6$. The setup presented in this technical report yields an average return of $74.5$ (best-performing policy out of 3 training runs with the cross-entropy loss with Treechop data incorporation). These results are obtained on a different texture set, since the competition textures are not publicly available.

\section{Discussion}

Imitation learning can work very well when sufficient demonstrations are provided. However, imitation learning has to deal with some of the same problems as reinforcement learning: the losses are weakly correlated with the actual performance of the policy at test time and the returns are unstable over time. The performance fluctuates a lot during training~(Figure \ref{fig:eval}). 

This distinguishes imitation learning from other supervised learning tasks, such as image classification on a fixed dataset. While single misclassified samples due to stochastic gradient descent have little influence on the overall performance of a standard classifier, on temporally correlated processes fluctuations can cause the neural network to predict a poor action for an essential early state. This can cause a performance collapse of the entire trajectory because future states where the network performs well are never reached.

Minecraft presents a challenging and exciting domain for sensorimotor learning. We hope that the strong imitation learning baseline described in this technical report can support future progress.

\bibliography{main}
\bibliographystyle{apalike}

\appendix

\clearpage
\section{Supplementary Material}
\beginsupplement

\subsection{Architectures}

The overall architecture of the neural networks consists of three parts. The first is a convolutional part that embeds the RGB image. If the environment also has a state vector, there is a fully-connected part that embeds the state vector. Finally, the last fully-connected part takes as input the image embedding concatenated with the state vector embedding if there is one.

The part that embeds the state vector consists of a single fully-connected layer of size $128$ except for Double DeepImpala where it is $256$.

As image embedding a convolutional body is used followed by one FC layer with $1024$ outputs ($2048$ for Double DeepImpala). 
Next we describe the different convolutional networks that were used to embed the input image.

\textbf{DQN:}
The Atari DQN-architecture \citep{mnih2015human}.

\textbf{Impala:}
There are three successive blocks where each block consists of a convolution followed, by max pooling, followed by two residual blocks. The channel sizes of the three blocks are $(
32, 64, 64)$ which is double the size of the original Impala architecture \citep{espeholt2018impala}.

\textbf{DeepImpala:}
The architecture~\cite{Obstacle} is different from the Impala architecture in two ways. First, the plain residual blocks are replaced with Fixup blocks \citep{zhang2019fixup}. Second, there are four blocks of (convolution $\rightarrow$ max pooling $\rightarrow$ Fixup block $\rightarrow$ Fixup block). The four blocks have channel sizes $(32, 64, 64, 64)$.

\textbf{Double DeepImpala:}
The image embedding is same as for DeepImpala except that the channel sizes are doubled to $(64, 128, 128, 128)$.

The last part consists of two fully-connected layers. The input is the image embedding concatenated with the state vector embedding. First a ReLU non-linearity is applied, followed by the first layer of size $1024$ except for Double DeepImpala where it is $2048$. Then another ReLU non-linearity is applied followed by the second fully-connected layer that has as many outputs as there are actions. If the network is trained with a cross-entropy loss, there is a further softmax applied in the end.

\subsection{Action Space}

The agent navigates through the Minecraft environment by a combination of the following basic movement sub-actions: \textit{forward}, \textit{backward}, \textit{left}, \textit{right}, \textit{jump}, \textit{sprint}, \textit{attack}, \textit{sneak}, and through continuous yaw and pitch control of the agent's camera orientation.
We discretized the camera movement into four actions of moving the camera by 22.5 degree in each direction: \textit{turnCameraUp}, \textit{turnCameraDown}, \textit{turnCameraLeft}, and \textit{turnCameraRight}. 
The camera actions of the human players in the MineRL dataset are given as continuous values and the camera is usually turned slowly. Therefore we select one of the four camera actions if in the next three states the camera movement in that direction exceeded 11.25 degrees. Also, during interaction with the environment a small Gaussian noise is added to the value of 22.5 to prevent the agent from getting stuck due to the discretization.

This results in 1280 discrete movement actions. To decrease the size, the for the tasks unimportant action \textit{sneak} and the combination of conflicting action pairs like \textit{forward} and \textit{backward} are removed.
Further, only one of the four camera movements is allowed per action and the total amount of basic actions is restricted to three sub-actions.
From all possible combinations with more than three sub-actions, they are removed in the following unimportant-to-important order until only three remain: \textit{sprint}, \textit{left}, \textit{right}, \textit{back}, \textit{turnCameraUp}, \textit{turnCameraDown}, \textit{turnCameraLeft}, \textit{turnCameraRight}, \textit{attack}, \textit{jump}, \textit{forward}. 

Otherwise all combinations of basic actions were considered, resulting in 112 individual movement actions. The remaining actions are for crafting, equipping and placing relevant items, leading to a total of 130 actions. As a last step we always set the \textit{jump} action to 1, unless the \textit{attack} action is used.

\subsection{State Space}

The state of the the agent consists of the first-person point-of-view RGB image and a vector with information about the inventory (except for the \textit{Treechop} task where the agent state consists of only the image).

The first four dimensions of the state vector are a one-hot encoding of the item the agent has in its mainhand. It is one of 
\{\textit{none}, \textit{wooden\_pickaxe}, \textit{stone\_pickaxe}, \textit{iron\_pickaxe}\}.
As the agent can carry items in different amounts in its inventory, for each item we either encode the number of times it is in the inventory as multi-hot if its average amount in the demonstration data is not too large and as a single float value otherwise. Here multi-hot means the amount of ones in a sub-vector is equal to the amount of the according item in the inventory. The size of the multi-hot vector for each item is given in Table \ref{tab:multi_hot_size}.
The amount of the items \textit{dirt},  \textit{cobblestone}, \textit{stone} in the inventory is represented by a single value which is the current amount of them devided by their average amount over the expert data.

\begin{table}
\centering
\caption{Size of the multi-hot encoding vector for each item.}
\begin{tabular}{lc}
\toprule
\textbf{Item} & \textbf{Size multi-hot vector} \\
\midrule
coal & 16 \\
crafting\_table & 3 \\
furnace & 3 \\
cobblestone & 16 \\
iron\_ingot & 8 \\
iron\_ore & 8 \\
iron\_pickaxe & 3 \\
log & 3 \\
planks & 64 \\
stick & 32 \\
stone\_pickaxe & 4 \\
torch & 16 \\
wooden\_pickaxe & 4 \\
\bottomrule
\end{tabular}
\label{tab:multi_hot_size}
\end{table}

\end{document}